\documentclass[conference]{IEEEtran}
\usepackage{cite}
\usepackage{amsmath,amssymb,amsfonts}
\usepackage{algorithmic}
\usepackage{graphicx}
\usepackage{textcomp}
\usepackage{xcolor}
\usepackage{subcaption}
\usepackage{multirow}
\def\BibTeX{{\rm B\kern-.05em{\sc i\kern-.025em b}\kern-.08em
    T\kern-.1667em\lower.7ex\hbox{E}\kern-.125emX}}
\begin{document}

\title{Improving Routability Prediction via NAS Using a Smooth One-shot Augmented Predictor
}

\author{\IEEEauthorblockN{1\textsuperscript{st} Arjun Sridhar}
\IEEEauthorblockA{\textit{CEI Lab} \\
\textit{Duke University}\\
Durham, USA \\
}
\and
\IEEEauthorblockN{2\textsuperscript{nd} Chen-Chia Chang}
\IEEEauthorblockA{\textit{CEI Lab} \\
\textit{Duke University}\\
Durham, USA \\
}
\and
\IEEEauthorblockN{3\textsuperscript{rd} Junyao Zhang}
\IEEEauthorblockA{\textit{CEI Lab} \\
\textit{Duke University}\\
Durham, USA \\
}
\and
\IEEEauthorblockN{4\textsuperscript{th} Yiran Chen}
\IEEEauthorblockA{\textit{CEI Lab} \\
\textit{Duke University}\\
Durham, USA \\
}
\and
% \and
% \IEEEauthorblockN{4\textsuperscript{th} Yiran Chen}
% \IEEEauthorblockA{\textit{ECE} \\
% \textit{Duke University}\\
% \and
% \and
}

\maketitle

\begin{abstract}
  Routability optimization in modern EDA tools has benefited greatly from using machine learning (ML) models. Constructing and optimizing the performance of ML models continues to be a challenge. Neural Architecture Search (NAS) serves as a tool to aid in the construction and improvement of these models. Traditional NAS techniques struggle to perform well on routability prediction as a result of two primary factors. First, the separation between the training objective and the search objective adds noise to the NAS process. Secondly, the increased variance of the search objective further complicates performing NAS. We craft a novel NAS technique, coined SOAP-NAS, to address these challenges through novel data augmentation techniques and a novel combination of one-shot and predictor-based NAS. Results show that our technique outperforms existing solutions by 40\% closer to the ideal performance measured by ROC-AUC (area under the receiver operating characteristic curve) in DRC hotspot detection. SOAPNet is able to achieve an ROC-AUC of 0.9802 and a query time of only 0.461 ms. 
\end{abstract}


\section{Introduction}
Traditional integrated circuit (IC) design is a complicated process involving many iterations of design and validation. 
% Several EDA tools have been developed to automate this process and flag potential issues. 
Routing, a step in this process, connects the components of an IC while avoiding DRC violations and ensures that the wires are not overcrowded. We focus on detailed routing (DR) hotspot detection, which is a vital step in the EDA design process. DR hotspot detection provides engineers with the location of hotspots in a proposed IC design. This allows the engineers to then change the placement of components to reduce the number of hotspots. Typically, this process is iterative with several rounds of modification and analysis. Traditional methods perform DR hotspot detection by first performing detailed routing, a computationally costly and time-consuming algorithm. Engineers are forced to wait several hours for results to generate, only to be informed of issues requiring a redesign. This problem has only worsened as the complexity of hotspot detection has significantly risen with the increasing complexity of ICs. As more and more components are able to be placed on a board, more wires are needed to connect these components\cite{routenet}. 

Recently, machine learning techniques have been employed using prior knowledge to improve DRC hotspot detection \cite{nas-craft}. Neural Networks (NNs) are able to perform hotspot detection without performing detailed routing. Thus, NNs are much faster than traditional methods, taking less than a second, as opposed the several hours of traditional methods. With rapid design feedback from ML models, engineers are able to develop high-quality solutions with fewer iterations than previously possible. Generative adversarial networks (GANs) proved to be a popular choice early on, but have been superseded by simpler and more efficient convolutional neural networks (CNNs) \cite{routenet, drc-cnn, pros,FPGA2_GAN, FPGA3_GAN}. NNs are traditionally used in a variety of image-based applications such as image recognition and semantic segmentation. DRC hotspot detection is very similar to semantic segmentation, as the goal is to determine which pixels of an image contain a specific attribute. In the case of hotspot detection, the attribute under consideration is the existence of hotspots present in an image of an IC layout. 

\begin{figure}[t]
    \centering
    \includegraphics[scale=0.87]{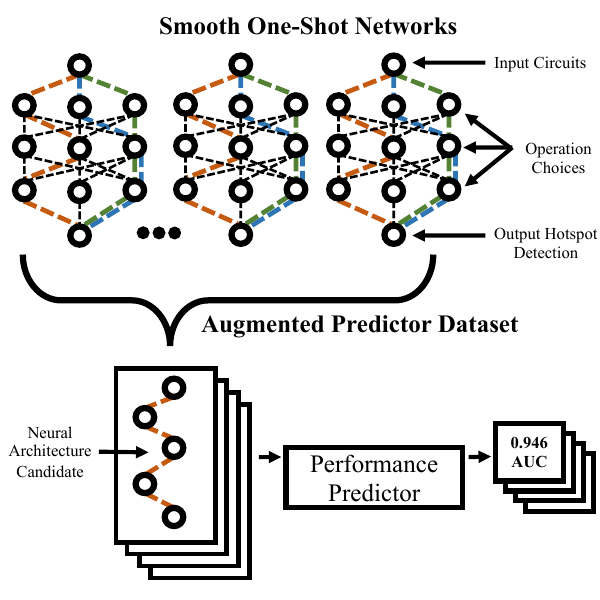}
    % \vspace*{-5pt}
    \caption{Several one-shot networks are independently trained to produce an augmented dataset for performance prediction. }
    % \vspace*{-10pt}
    % \Description{Flow chart of algorithm overview.}
    \label{fig:overview}
\end{figure}
% \vspace*{-1pt}
CNN models pose a design challenge of their own as the structure of a neural network greatly affects its performance. Automated Machine Learning (AutoML), particularly Neural Architecture Search (NAS), empowers design automation for a wide array of ML models with minimal to no human intervention. The field of NAS serves to ease the burden of developing CNNs and holds many techniques and tricks. Models generated through NAS have surpassed the performance of state-of-the-art manual designs, exhibiting significantly improved model accuracy and computational efficiency. 
% To achieve this, NAS initially defines an architecture search space\cite{lissnas} and then employs specific search strategies, such as reinforcement-learning-based\cite{zoph2017neural} or evolutionary-guided methods\cite{real2017large}, to intelligently discover promising architectures for a given task, such as image segmentation. 

% One-shot NAS utilizes weight sharing in the form of a one-shot network to significantly improve NAS efficiency and quality. This technique utilizes a one-shot network that employs a shared set of weights to train multiple candidates simultaneously. While this approach substantially reduced training time, it had the drawback of restricting architecture exploration, leading to a potential impact on performance. To overcome this limitation, OFA \cite{cai2019ofa}, Attentive NAS\cite{wang2020attentivenas}, and various other research works adopted novel path sampling methods and a range of optimization techniques to train these networks. As a result, one-shot networks demonstrated remarkable efficiency and achieved state-of-the-art NAS performances across a diverse range of tasks. 
Applying modern NAS methods directly to DRC hotspot detection proved challenging in our initial findings. The primary quality metric used in hotspot detection is ROC-AUC, which is a by-product of the network accuracy. While the ROC-AUC and accuracy are linked, there is a significantly higher variance in ROC-AUC. Since ROC-AUC is not the primary metric optimized during network training (network accuracy), using a one-shot network to directly search for an architecture is not possible. The loss landscape of a one-shot network is smooth with respect to accuracy, unlike ROC-AUC. To alleviate and address these issue we propose a NAS methodology for DRC hotspot detection coined SOAP-NAS, which utilizes a novel combination of one-shot and predictor-based NAS to produce the final architecture and novel data augmentation. Figure \ref{fig:overview} provides a simplified overview of the proposed algorithm. In addition, search space selection proves to be an important factor for NAS performance. Through optimization for hotspot detection, we are able to further improve performance. Our main contributions can be summarized as follows: 
\begin{itemize}
    \item We empirically determine the best NAS search space for DRC hotspot detection through comparing performance on 3 popular NAS search spaces.
    \item We propose a novel k-shot smoothing technique to combat the training-related variance of the target metric through training many one-shot networks.
    \item We propose a novel data augmentation method for predictor-based NAS to address the shift in the optimization objective from network accuracy to ROC-AUC and to further capture the increased variance.
    \item We combine these techniques into an end-to-end NAS algorithm that produces state-of-the-art results for DRC hotspot detection in performance and speed.
\end{itemize}
We achieve state-of-the-art results with SOAP-NAS achieving 40\% closer to ideal performance when compared with existing approaches. The final model produces hotspot detection with 0.9802 ROC-AUC.
\section{Background}
\subsection{Routability Prediction using Machine Learning}
Hotspot problems in the EDA domain primarily refer to potential manufacturing defects in the design of integrated circuits (ICs) that can lead to functional failures or lower yield \cite{hotspot1, hotspot2, zhang2024qplacer}. The downsizing of transistor dimensions has amplified the importance of hotspot detection. The complexity of these problems often necessitates significant computational resources, particularly for intricate IC designs. The EDA community actively explores diverse methods that empower designers to preemptively circumvent DRC violations, such as leveraging machine learning to predict the congestion count \cite{cc_sv, cc_sv2} or identify the location of these hotspots \cite{FPGA4, pin, wang2022lhnnlatticehypergraphneural, 1e9716e0773d44d2b45ca4038666525b}. 

CNNs have been commonly employed for predicting the number of violations \cite{drc-cnn, FPGA}. Simultaneously, Fully Convolutional Networks (FCNs) have gained prominence for their efficacy in DRC hotspot detection, capitalizing on their strength in identifying pixel-wise properties in a two-dimensional layout \cite{routenet, pros}. Several studies reformulated the problem into image generation tasks, employing FCN and conditional GAN to build correlations \cite{FPGA2_GAN, FPGA3_GAN}. However, earlier work heavily relies on manual design \cite{routenet, drc-cnn, pros}, requiring both ML and EDA expertise. This can lead to extended model development time. These models, structured hierarchically with limited branches, inspired us to enhance the search space.  We aim
for improved flexibility and more branching, thus facilitating
notably distinct model structures enabling better performance.

\subsubsection{ROC-AUC}
ROC-AUC serves as the metric for evaluating model performance on DRC hotspot detection. ML models trade accuracy for speed by skipping full algorithmic detailed routing. While much faster, NNs can erroneously report hotspots that don't exist (FPR) or fail to report hotspots that do exist (TPR). As such, ROC-AUC is the chosen metric to capture the trade-off between true positive and false positive rate. Through ROC-AUC, we can accurately evaluate and compare the performance of ML-accelerated hotspot detection. An ROC-AUC performance of 1.0 would indicate that the model correctly classifies all hotspots with no false positives. Engineers are able to tolerate the error rate of CNNs in exchange for the rapid reduction in runtime. However, a model that produces excessive false positives can cause engineers to spend time fixing issues that are unnecessary. 
\subsection{Neural Architecture Search}
The task of designing a CNN presents researchers with a challenging endeavor as they strive to optimize multiple constraints, including latency, model size, and accuracy. To address this challenge, recent efforts in NAS aim to automate the process of creating and identifying CNN architectures that fulfill these constraints.

NAS can be broken down into two main components: the search space and the NAS method. The search space defines the set of candidate architectures being considered by a NAS method. The search space can also be thought of as a family of models. Typically, search spaces contain a set of operations and a set of rules defining how these operations are connected. Modern search spaces vary greatly in size and complexity. Search space shrinkage \cite{lissnas}, a recent field of development, demonstrated the importance of search space selection. The primary objective of NAS methods is to predict the accuracy of a candidate neural architecture. In 2017, the concept of NAS with Reinforcement Learning was introduced, utilizing an RL-based controller that iteratively proposed and trained candidate architectures \cite{zoph2017neural}. However, the training of both the controller and candidates incurred significant computational costs. As a result, various alternative approaches have been explored, including the utilization of genetic algorithms \cite{real2017large}, Bayesian optimization \cite{pmlr-v28-bergstra13}, and predictors \cite{wen2020neural}, to overcome these computational complexities. The introduction of weight sharing through one-shot NAS served as the next major breakthrough \cite{guo2019singlepath}. One-shot networks proved to be efficient and resulted in improved performance across many tasks. Our work uses one-shot as the backbone NAS method in combination with a gradient boost predictor to determine the best architecture for DRC hotspot detection.
% Current NAS approaches are still costly to run and require some level of human expertise to carefully design and construct the search space.
One-shot approaches suffer from challenges resulting from the complexity of representing many candidate architectures often numbering in excess of $10^{6}$. These approaches are also subject to quality loss when the target metric changes from the metric during training and have a high degree of variance across the space. The combination of these problems highlights the significance of our work. 

\section{Methodology}

\begin{table}[t]
\centering
\footnotesize
\caption{NAS performance on EDA demonstrated across popular search spaces using one-shot NAS}
\label{table:nas-prelim}
\begin{tabular}{l|c|c|c}
\hline
Search Space  & Pearson Correlation & Kendal Tau & AUC \\ \hline
NASBench101   & 0.52                & 0.45       & 0.95    \\ \hline
ShuffleNetV2  & 0.36                & 0.32       & 0.939    \\ \hline
TransNASBench & 0.23                & 0.09       & 0.941   
\end{tabular}
     \vspace{-5mm}
\end{table}
% need a reason/insights as to why NASBench101 is the best SS
% We expect to see signficantly higher values for Pearson Correlation and Kendal Tau meaning that the OneShot networks are not very good at predicting AUC. In addition, we would also expect to see an improvement in AUC when compared with previous works which is currently lacking.
\subsection{Search Spaces}
We consider 3 search spaces: NASBench101 \cite{ying2019nasbench}, ShuffleNetV2 \cite{zhang2017shufflenet}, and TransNASBench \cite{transNASBench}. These search spaces represent popular CNN families for a variety of image-based tasks. TransNASBench \cite{transNASBench} contains networks evaluated across a domain of various image-based tasks ranging from semantic segmentation to object recognition and includes 7 diverse tasks. While NASBench101 \cite{ying2019nasbench} and ShuffleNetV2 \cite{zhang2017shufflenet} are designed for image recognition, they appear to perform well for DRC Hotspot detection. NASBench101 contains 432k unique architectures using a backbone of cells stacked in a repeated fashion. Each cell contains 5 nodes and up to 7 edges with each node containing 1 of 3 operation choices. ShuffleNetV2 contains $4^{20}$ unique architectures through 20 layers stacked with 4 operations choices at each layer. TransNASBench encoporates a bilevel macro and micro level search space utilizing a cell and backbone structure the details of which are outlined in that work \cite{transNASBench}.

\subsection{Best Search Space for DRC Hotspot Detection}
We begin by exploring the use of one-shot NAS on DRC hotspot detection. From Table \ref{table:nas-prelim}, we observe the Pearson correlation, Kendal Tau, and best queried AUC from the one-shot network. These AUC results indicate that the CNN architecture designs perform well at the DRC hotspot detection tasks. Pearson correlation and Kendall Tau are ranking correlation metrics that represent the degree to which the one-shot network is able to predict the ground truth ROC-AUC for networks in the search space. One-shot NAS struggles to produce good Pearson correlation and Kendall Tau when searching for ROC-AUC as seen in Table \ref{table:nas-prelim}. Two main factors could be attributed to these results. Firstly, the one-shot network is trained using accuracy as the metric that feeds into the loss function when updating weights. The resulting one-shot network has a relatively smooth loss landscape meaning that networks that are close together have similar accuracy and loss values. While ROC-AUC generally mirrors accuracy, there are cases where this relation breaks down. This results in degraded performance when querying the one-shot network for a different metric than the metric used during training. Secondly, ROC-AUC as a metric has higher variance than accuracy. The variance might be due to the fact that AUC is not directly linked to accuracy. For example,  a model with high accuracy might also have a higher false positive rate leading to a flatter ROC and lower overall ROC-AUC. The increased variance could directly contribute to the poor correlation metrics.
% In this work, accuracy refers to ROC-AUC interchangeably for the purpose of NAS metrics such as Pearson Correlation and Kendal Tau.

NASBench101 contains architectures with the highest ROC-AUC indicating the best performance for routability prediction. NASBench101 also has the highest values for one-shot performance as indicated by Pearson correlation and Kendall Tau. We determined the NASBench101 search space to be the best fit for applying one-shot NAS to DRC hotspot detection. However, the overall correlation of accuracies could be improved as shown in the sections that follow. 

% How NAS results apply to EDA? what can we learn from the NAS results? what type of architectures perform well on EDA.

% Figure with eda input and output to show difference between existing methods and new method (check with Junyao )

% Details of NAS should be covered concretely 

\subsection{Variance in ROC-AUC}
\begin{figure}[t]
    \centering
    \includegraphics[scale=1]{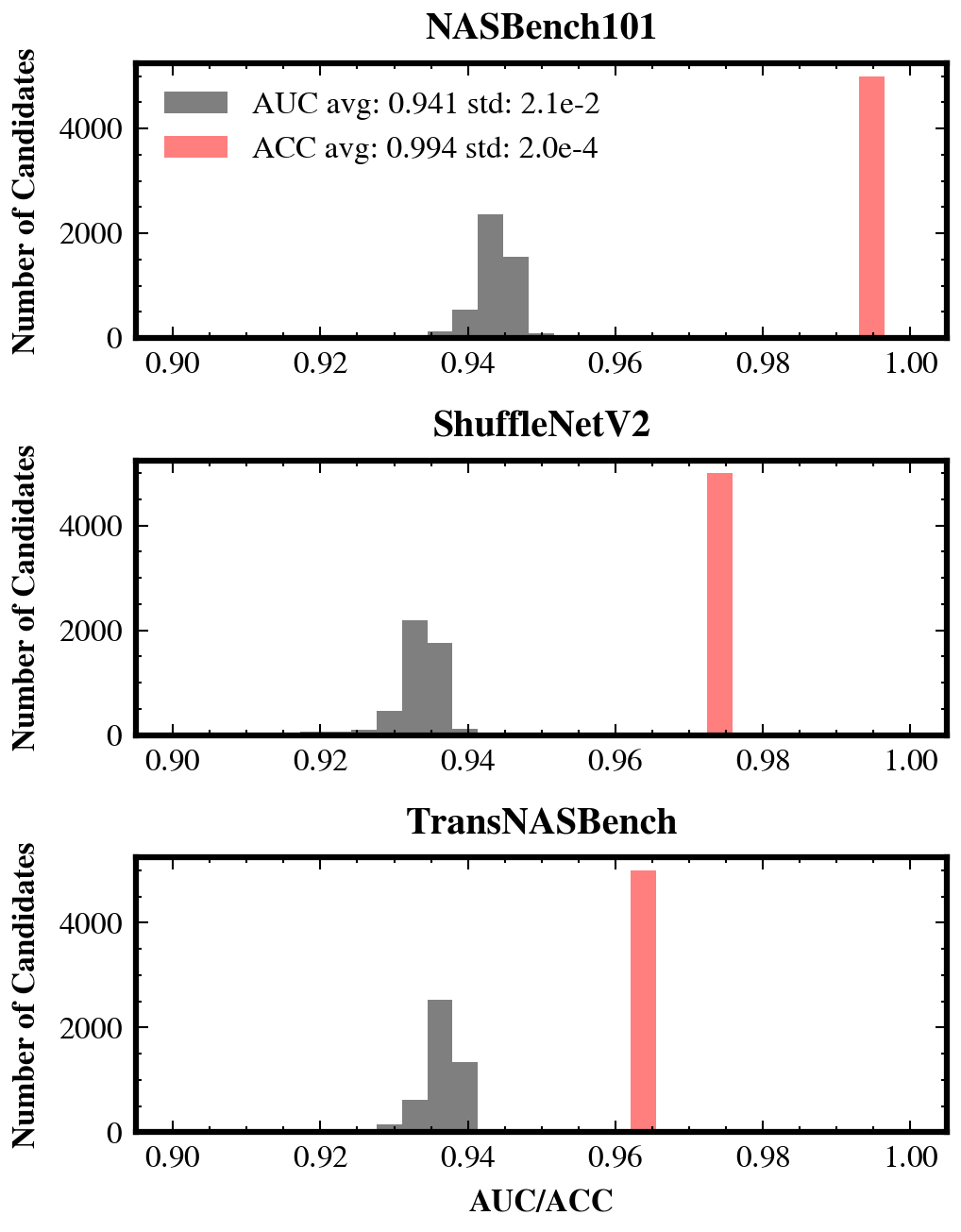}
    \caption{Histogram of queried AUC and ACC}
    % \vspace*{-10pt}
    % \Description{Histogram of AUC and ACC.}
    \label{fig:histo}
\end{figure}

\begin{table}[b]
\begin{center}
\caption{Mean and Variance for stand alone trained architectures sampled from the NASBench101 search space.}
\label{table:var}
\begin{tabular}{l|l|l}
\hline
Metric   & Mean & Variance \\ \hline
ROC-AUC  & 0.96 & 0.0144   \\ \hline
Accuracy & 0.99 & 0.0043 
% \vspace*{-10pt}
\end{tabular}
\end{center}
\end{table}
The NASBench101 search space in combination with one-shot NAS produces favorable results, but there is still much room for improvement in correlation. 
% Initially, one may think that the performance difference could be due to the change in target metric from network accuracy to ROC-AUC. 
When looking at histogram distributions of AUC compared to accuracy in Figure \ref{fig:histo}, we observe that AUC has significantly greater spread and higher standard deviation. This makes final performance more prone to randomness in training and sampling in the one-shot network. We also found that the variance of AUC is higher than that of accuracy, as shown in Table \ref{table:var}. This table is generated by sampling 50 networks from the NASBench101 search space and training them from scratch for 20 epochs. These 50 network are then retrained from scratch 10 times to get averages for mean and variance of ROC-AUC and accuracy. From these results, we can isolate that the increased variance is a property of the task and corresponding ROC-AUC metric and not due to the training process of the one-shot network or its related hyperparameters. For a one-shot network to perform well, we must capture this increased variance. 
% This table was produced by training 100 networks 5 times from scratch for 15 epochs to determine the variance in both accuracy and AUC. 
% While this may have less of an impact when the networks are trained to completion 
% This metric gives a good benchmark into how prone the networks are to randomness in training and weight initialization. 
% We found that the variance in AUC is significantly higher than accuracy when networks are trained in a standalone fashion. 

\subsection{One-shot Training}
NAS begins with training the one-shot network. The one-shot network is trained for 240 epochs using a learning rate of 0.02 and a training batch size of 32. We use a ghost batch size of 8 paired with a random path sampling strategy for each of the search spaces. Once the supernetwork is trained, we then generate a validation set of (network, AUC) pairs. This dataset is then used to develop a performance predictor that given a candidate architecture will return the estimated AUC of that network. We perform a train-test split with cross-validation and determined that the XGBoost predictor performs well. To validate the performance in an end-to-end fashion, we also train several candidate architectures from scratch in a stand-alone fashion. This is done to ensure AUC of the candidate network matches the predicted AUC. 

Pearson correlations of AUCs in Table \ref{table:nas-prelim} are unexpectedly low. 
% There could be several potential causes. First, we determine if the correlation issue is due to some property of the objective change from accuracy. To do this, we recalculate the correlation of accuracies. However, we see roughly the same correlation. 
We determined that the cause of the low correlation is due to the variance in final metrics to the random initialization of weights.

\begin{figure}[t]
    \centering
    \includegraphics[scale=0.50]{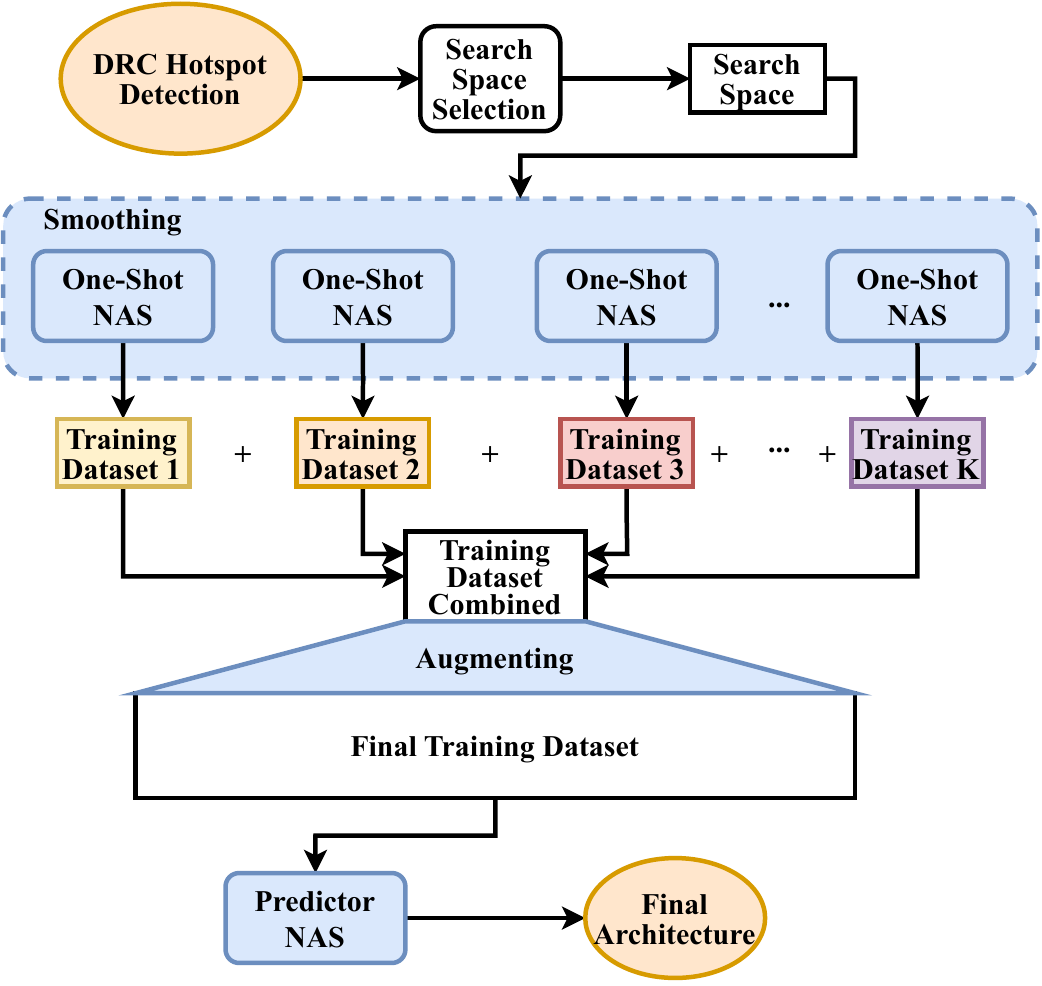}
    \caption{Overview of SOAP-NAS}
    % \vspace*{-10pt}
    % \Description{Overview of SOAP-NAS.}
    \label{fig:soap}
\end{figure}

\subsection{Smoothing}
We employed a novel smoothing technique to alleviate the issue of increased variance. To generate the dataset used to train the predictor, we train several one-shot networks from scratch. The initialization and path sampling randomness ensures that these one-shot networks are different and cover various training scenarios. Through experimentation, we found that training 5 one-shot networks provided sufficient diversity and greatly improved the correlation metrics. There was marginal improvement of results when training more than 5 networks. The details of this can be found in the ablation study section below. We then build the dataset for the predictor by querying each of the one-shot networks for a set of candidates, keeping the top performance value of networks appearing in two or more sets. The resulting dataset helps to mitigate the impact of randomness in training and initialization of the networks. While this gets us closer to the desired ideal result, there is still noise introduced in each of the one-shot networks' training procedure. In addition, the final dataset size is not large enough to effectively train a predictor-based NAS method. 
\subsection{Augmenting}
To efficiently generate enough training data for the predictor, we augment the training data by sampling values in the combined dataset and adding artificially generated noise. The noise added to the models follows the underlying variance of the ROC-AUC, which is estimated through training several standalone networks from scratch earlier. Again, through experimentation, we found that augmenting by $7x$ serves sufficient. The details of this can be found in the ablation study section below. Now the training data for the predictor closely resembles the ground truth data acquired from training the candidate architectures from scratch. The predictor now has a much better Pearson Correlation of 0.65 when comparing the predicted AUC against the best stand-alone trained AUC of an independent validation set. 
\subsection{SOAP-NAS}
Figure \ref{fig:soap} provides a detailed overview of SOAP-NAS. We begin by selecting the best search space for DRC hotspot detection. This search space is then used to train $k$ one-shot networks in order to capture the increased variance of ROC-AUC in the "smoothing" step. By training several one-shot networks, we are able to mitigate the training biases that can skew the ROC-AUC metric. We found that $k=5$ provides sufficient results in the ablation study section below. Each of the datasets produced by the one-shot networks are then combined and augmented to produce the dataset used to train the performance predictor. Augmentation is done through sampling the datasets and augmenting with a noise model obtained through fitting to (network, AUC) pairs trained from scratch. This is done to efficiently increase the size of the dataset by $7x$ and to further capture the increased variance of ROC-AUC. Finally, predictor-based NAS is performed to extract the final architecture. We call this new technique SOAP-NAS, Smooth One-shot Augmented Predictor NAS. Note we only need to run SOAP-NAS once to produce a neural candidate that can perform routing on many different circuit designs.
\section{Evaluation}
\subsection{Experiment Setup}
\subsubsection{Dataset}
A comprehensive dataset, designed to facilitate extensive comparative evaluation, was compiled from 74 designs across multiple benchmarks including ISCAS’89 \cite{ISCAS}, ITC’99 \cite{ITC}, IWLS’05 \cite{IWLS}, and ISPD’15 \cite{ISPD}. To augment the dataset, various placement solutions were created for each design using diverse logic synthesis or physical design settings. Thus, totaling 7,000 placement solutions. Logic synthesis was executed using Design Compiler, while Innovus facilitated physical design with the NanGate 45nm technology library. Input feature maps were collected from post-placement stage, with ground-truth DRC results obtained after post-detailed routing.

\subsubsection{Baselines}
The following schemes are implemented and assessed in evaluations
\begin{itemize}
\item RouteNet \cite{routenet} is the first to employ fully convolutional network (FCN) for forecasting DRC hotspots based on cell placement and global routing data.
\item PROS \cite{pros} is another technique using FCN to model the correlation between congestion locations during global routing and cell placement.
\item NAS-crafted \cite{nas-craft} is the optimal model from its search space, using NAS to automate the development of high-quality neural architectures for routability prediction and guiding cell placement towards more routable solutions.
\item SOAPNet uses a one-shot network to generate a dataset used to train a predictor on (network, AUC) pairs. The dataset is smoothed and augmented to closely resemble actual candidate architectures.
\end{itemize}

\begin{table}[t]
\caption{Routability predictors performance on 7k placement dataset \textit{*reproduced results}}
\label{table:auc}
\begin{center}
 \begin{tabular}{l|c|c}
\hline
Models   & ROC-AUC  & Query Time in MS\\ \hline
RouteNet* & 0.9511  & 0.476 \\ \hline
PROS*     & 0.9580  & 0.591 \\ \hline
cGAN*     & 0.9337  & 0.631 \\ \hline
NAS-Crafted*  &  0.9627 &  0.521\\ \hline
\bf{SOAPNet}   & \bf{0.9802} & \bf{0.461}  \\ 
\end{tabular}   
\end{center}
\vspace{-3mm}
\end{table}

\begin{table}[t]
\caption{Comparison of the DRC hotspot detection}
\label{table:compl}
\begin{tabular}{l|cccc|c}
\hline
\multicolumn{1}{c|}{\multirow{2}{*}{Models}} & \multicolumn{4}{c|}{ROC\_AUC on designs}                                                                                                   & \multirow{2}{*}{\begin{tabular}[c]{@{}c@{}}ROC-AUC \\ on all 74\end{tabular}} \\ \cline{2-5}
\multicolumn{1}{c|}{}                        & \multicolumn{1}{l|}{s349}           & \multicolumn{1}{l|}{mem\_ctrl}      & \multicolumn{1}{l|}{b17}            & \multicolumn{1}{l|}{DSP} &                                                                               \\ \hline
RouteNet                                     & \multicolumn{1}{c|}{0.829}          & \multicolumn{1}{c|}{0.844}          & \multicolumn{1}{c|}{0.902}          & 0.866                    & 0.847                                                                         \\ \hline
PROS                                         & \multicolumn{1}{c|}{0.487}          & \multicolumn{1}{c|}{0.483}          & \multicolumn{1}{c|}{0.478}          & 0.489                    & 0.676                                                                         \\ \hline
cGAN                                         & \multicolumn{1}{c|}{0.516}          & \multicolumn{1}{c|}{0.515}          & \multicolumn{1}{c|}{0.521}          & 0.517                    & 0.510                                                                         \\ \hline
NAS-crafted                                  & \multicolumn{1}{c|}{0.865}          & \multicolumn{1}{c|}{0.891}          & \multicolumn{1}{c|}{0.911}          & 0.884                    & 0.865                                                                         \\ \hline
\textbf{SOAP-NAS}                            & \multicolumn{1}{c|}{\textbf{0.889}} & \multicolumn{1}{c|}{\textbf{0.902}} & \multicolumn{1}{c|}{\textbf{0.930}} & \textbf{0.897}           & \textbf{0.901}                                                               
\end{tabular}
\end{table}

\begin{figure}[t]
    \centering
    \includegraphics[scale=1]{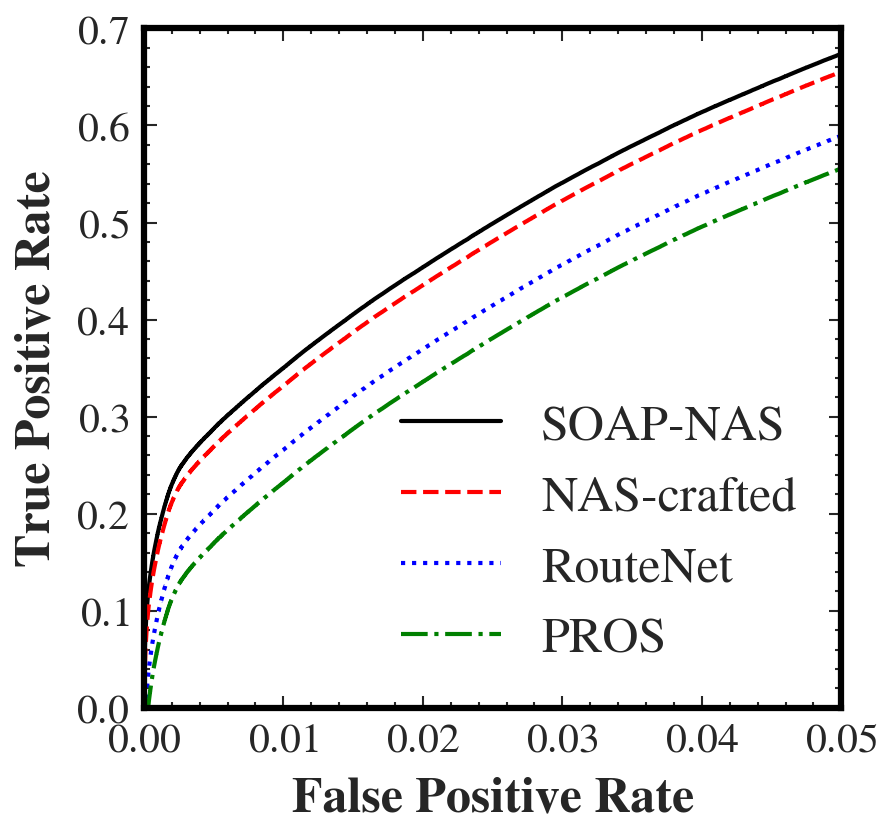}
    \caption{ROC curves show the superiority of SOAPNet compared to existing works. Curves do not intersect and viewing area is cropped to relevant region.}
    % \Description{ROC curves for SOAPNet.}
    \label{fig:edf}
    % \vspace{-5mm}
\end{figure}

% \begin{table}[t]
% \caption{Query time in milliseconds on average for DRC Hotspot Detection}
% \label{table:timeing}
% \begin{center}
% \begin{tabular}{l|c}
% \hline
% Method      & Query Time in MS \\ \hline
% RouteNet    & 0.476            \\ \hline
% PROS        & 0.591            \\ \hline
% NAS-Crafted & 0.521            \\ \hline
% \bf{SOAPNet}    & \bf{0.461}           
% \end{tabular}
% \end{center}
% % \vspace{-3mm}
% \end{table}

\subsection{DRC Hotspot Detection Evaluation Results}
The comparison of ROC-AUC values are presented in Table \ref{table:auc}. The best model produced by SOAPNet achieves an impressive AUC of 0.9802 among all evaluated 7k evaluated placements. 
% The results surpasses two human-crafted baselines, PROS and RouteNet, by 0.022 and 0.0291 in absolute value, respectively. 
Remarkably, although another NAS-oriented baseline provides higher AUC values when compared with human-crafted baselines, its result is still far from the value achieved by SOAPNet. While an improvement of 0.02 in AUC might seem marginal, the model produced by SOAPNet approaches the ideal ROC-AUC of 1.0 by 40\%, when compared with the best model of NAS-crafted baseline. 
These results demonstrate the superiority of SOAPNet in hotspot detection and its potential for further advancements in DRC optimization. 
Table \ref{table:compl} shows the direct comparison of results with the previous work. Prior works utilized the standard 74 design dataset, which is limited in scope and may not exhaustively test the models. Hence, our 7k placement dataset should serve as a new standard for routablity prediction.

% The ROC curves are illustrated in Fig.\ref{fig:edf}, which correspond to the ROC-AUC values in Table \ref{table:auc}. 
Then, we illustrate the ROC curves in Fig.\ref{fig:edf}.
The curves are truncated to show the difference in performance between SOAP-NAS and prior methods. The curves do not intersect outside the viewing region. When comparing the true positive rate of the SOAPNet model (y-axis) with PROS at the same false alarms cost (x-axis), SOAPNet exhibits notably higher performance. Although NAS-crafted and RouteNet show similar trends in the diagram, their true positive rates remain lower than SOAPNet. 

Table \ref{table:auc} presents the query time for DRC hotspot detection in four schemes. SOAPNet consistently outperforms the baselines, reducing the required time by ranging from 3\% to 28\%. Possessing with the better inference efficiency, the architecture produced by SOAPNet 
can serve as a superior choice for efficient placement optimization that usually requires multiple models queries.
% has simplicity feature that can improve the efficiency of evaluation. \JZ{[Not sure it is a correct discussion to express the benefit in query time]}

The comparative analysis of previous works, PROS and RouteNet, reveals lower hotspot detection accuracy on many designs. This is accompanied by longer query times during evaluation, despite our extensive optimization efforts and hyperparameter tuning. The underlying cause for these performance disparities could potentially stem from the inherent over-complexity of their architectures. Consequently, their models might face challenges in handling the diverse data characteristics present across training and testing designs from different benchmarks, which sets them apart from other baselines. 
% Furthermore, NAS-crafted model demonstrates its performance over the human-crafted techniques, but it still falls short in terms of ROC-AUC and query time, contrasting with SOAPNet. 
% This indicates the effectiveness of the novel NAS method, smoothing and augmenting the dataset, used to build the predictor. 
% \vspace{-2mm}
\subsection{Improvement from Prior Methods}
\begin{figure}[t]
    \centering
    \includegraphics[scale=0.42]{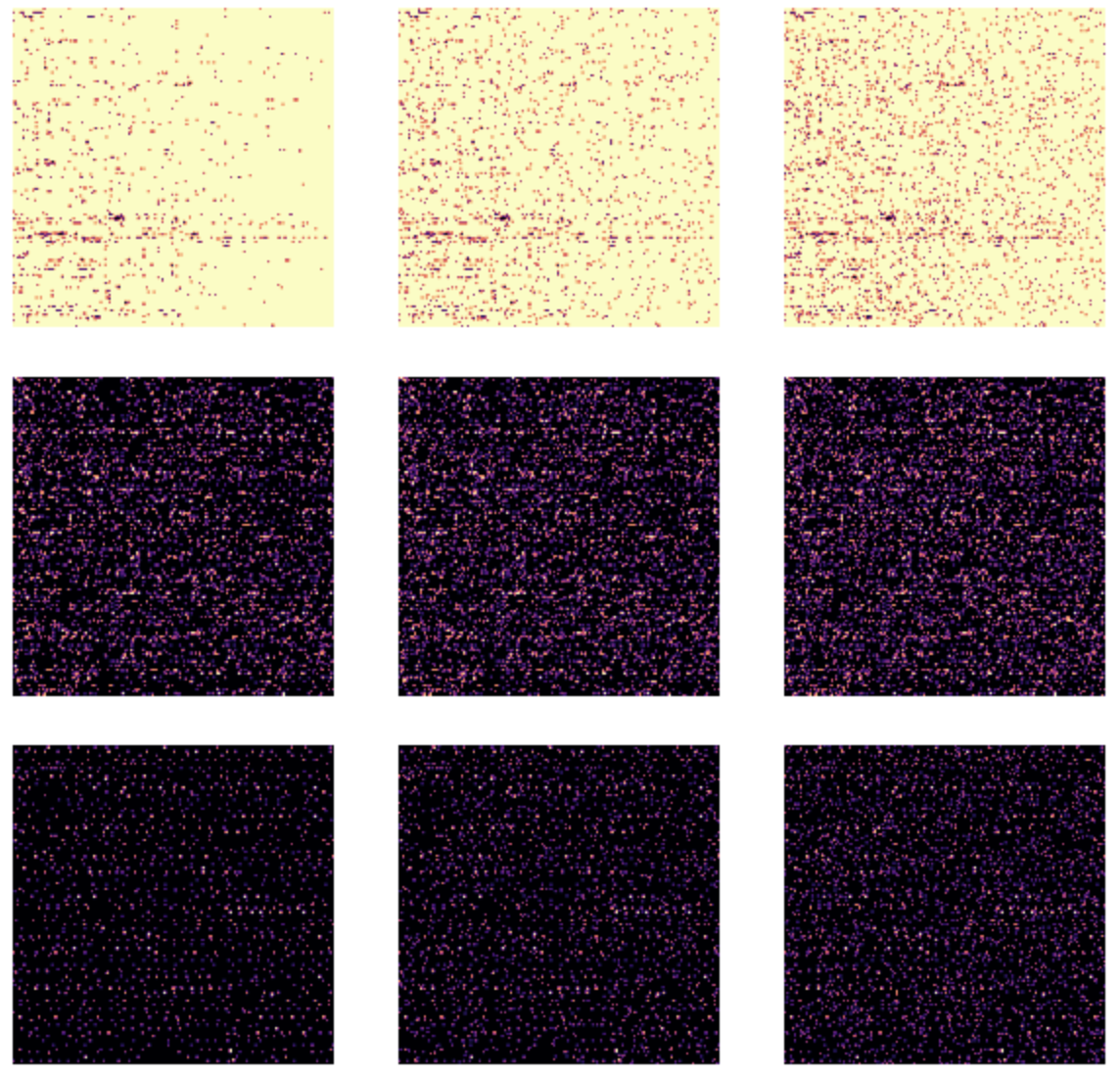}
    \caption{Left column shows examples of hot spot labels at different densities. The middle column is the output labels of our work and the right column is the best output from previous works.}
    % \Description{Effect of improvement.}
    \label{fig:improv}
    % \vspace{-7mm}
\end{figure}
In contrast to the previous state-of-the-art methodology, our study demonstrates a notable improvement in ROC-AUC by achieving 40\% closer to the ideal ROC-AUC. Despite the seemingly marginal disparity between AUC values of 0.98 and 0.96, the visual representation in Figure \ref{fig:improv} elucidates the substantive nature of this advancement. The middle column showcases the annotated outputs for hotspot identification generated by SOAPNet, illustrating a striking resemblance to the ground truth images presented in the leftmost column. Conversely, the rightmost column exhibits the annotated outputs corresponding to an ROC-AUC equivalent to that achieved by NAS-crafted. The depicted top row illustrates a circuit manifesting a substantial number of Design Rule Check (DRC) violations, while the middle row portrays a circuit with a moderate frequency of violations. Lastly, the bottom row delineates a circuit characterized by minimal DRC violations. This comparative analysis vividly underscores the superior efficacy of our algorithm across scenarios characterized by varying degrees of violation occurrences. Although less conspicuously discernible, a significant enhancement in label accuracy persists even in scenarios featuring a moderate level of DRC violations.

Our approach represents a significant leap forward from prior research endeavors. By employing techniques such as smoothing and augmentation, SOAP-NAS markedly surpasses the performance of previous methods. While NAS-crafted initiated the integration of NAS into DRC hotspot detection, its approach left ample room for advancement. NAS-crafted merely applies off-the-shelf NAS algorithms.
%to enhance the design of an FCN for hotspot violation detection, lacking specificity to address the intricacies of DRC hotspot detection. 
Our proposed SOAP-NAS methodology is tailored to directly tackle the challenges posed by DRC hotspot detection. Moreover, when juxtaposed with earlier methodologies like RouteNet and PROS, the degree of improvement becomes even more pronounced. This advancement in generating superior models for predicting DRC violations holds significant promise for enhancing the efficiency and reliability of machine learning-accelerated development tools.

\subsection{Ablation Study}

\begin{figure}[t]
    \centering
    \includegraphics[scale=0.68]{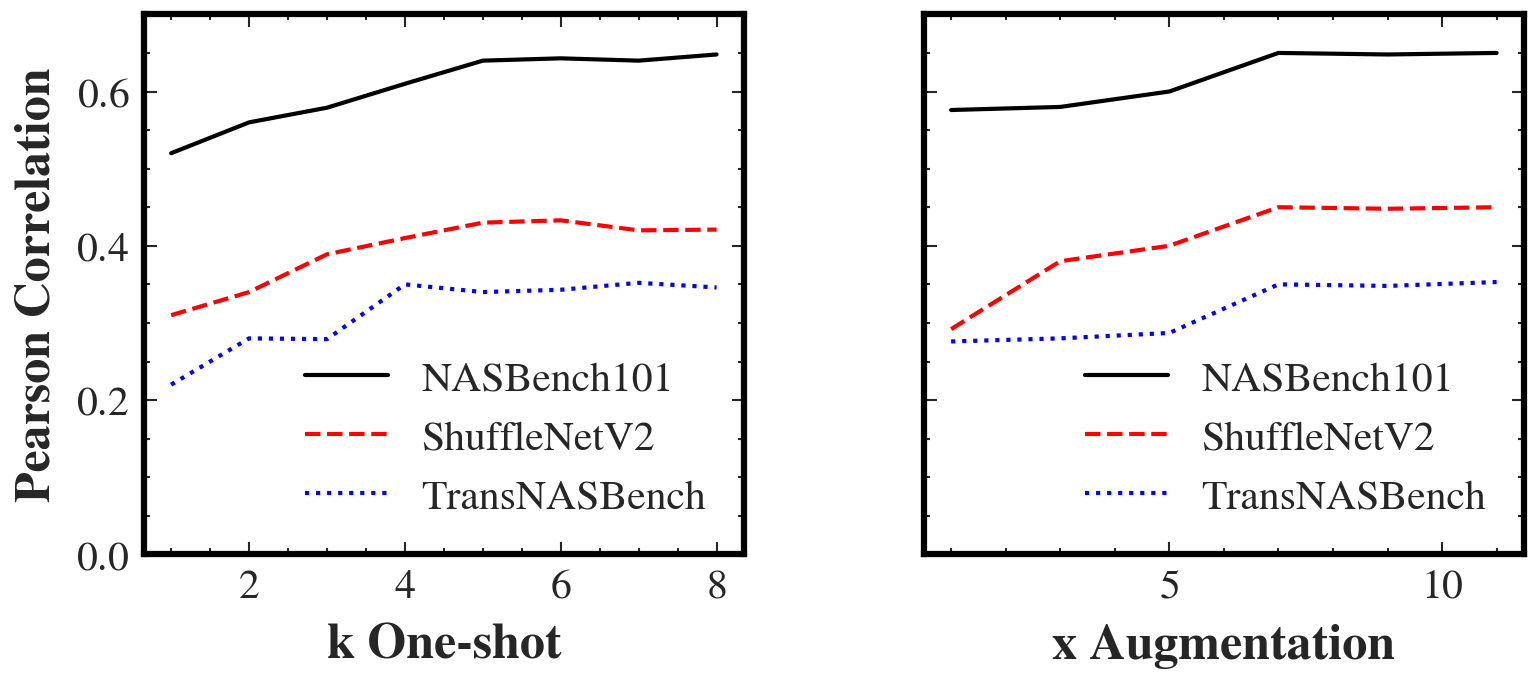}
    \caption{The effect of smoothing and augmenting data across 3 search spaces}
    % \Description{Ablation Study.}
    \label{fig:abla1}
    % \vspace{-5mm}
\end{figure}

\begin{figure}[t]
     \centering
     \begin{subfigure}[b]{0.24\textwidth}
         \centering
         \includegraphics[width=\textwidth]{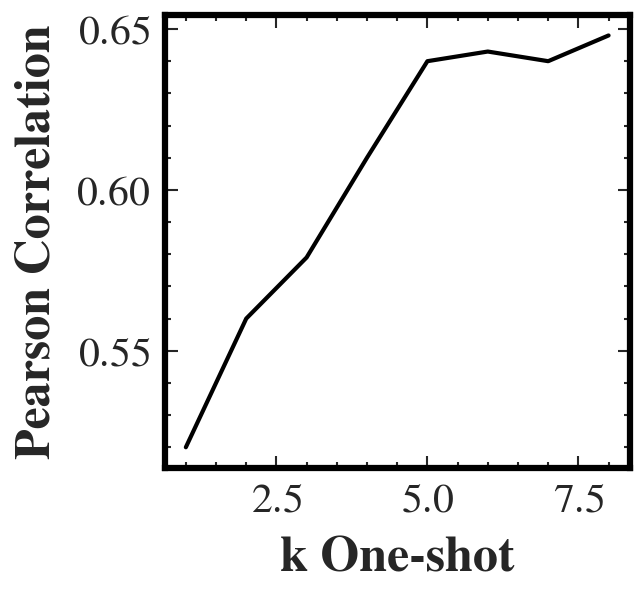}
         % \caption{ }
         % \label{fig:nas-k}
     \end{subfigure}
     \hfill
     \begin{subfigure}[b]{0.24\textwidth}
         \centering
         \includegraphics[width=\textwidth]{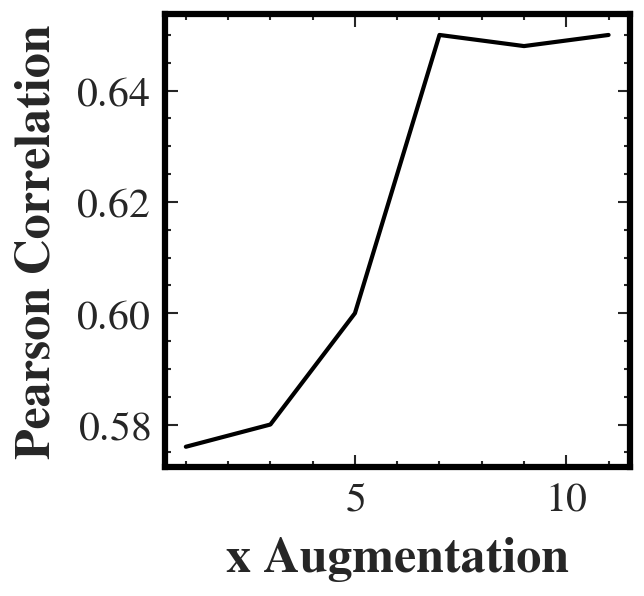}
         % \caption{ }
         % \label{fig:nas-x}
     \end{subfigure}
     \caption{Importance of smoothing and augmenting data on NASBench101}
     % \Description{Hyperparameter tuning graphs.}
     \label{fig:abla}
     % \vspace{-7mm}
\end{figure}

We further demonstrate the effectiveness of our technique through an ablation study. We found that removing SOAP-NAS results in significant performance loss. SOAP-NAS proves to be an effective technique to perform NAS for DRC hotspot detection due to the change in the target metric from the trainable metric of accuracy and the increased variance that this new metric brings. Figure \ref{fig:abla1} and \ref{fig:abla} show the impact and importance of smoothing and augmenting the data. 
In Figure \ref{fig:abla1}, we observe that the explorations of 3 search spaces have the same trend (plateauing at $k=5$ and $x=7$ respectively). 
Specifically, using 5 one-shot networks provides a good improvement in the Pearson Correlation of queried ROC-AUC to standalone ground truth ROC-AUC. Using a data augmentation factor of 7 also shows significant improvement in correlation. These two values serve as important hyperparameters of SOAP-NAS. Across both parameters, we see significant improvement through the addition of smoothing and augmentation as indicated by the increase in correlation from 1 to 5 and 1 to 7 respectively. Figure \ref{fig:abla} demonstrates the effect of smoothing using $k$ one-shot networks as $k$ ranges from 1 to 8. We observe that the graph shows marginal improvement after $k=5$ and thus choose this as the number of one-shot networks to train in SOAP-NAS. Furthermore, we can see the effect of not smoothing, represented by $k=1$, produces a poor Pearson correlation, demonstrating the importance of smoothing in the first place. Figure \ref{fig:abla} shows the effect of augmenting the data used to train the predictor. Similar to smoothing, data augmentation improves performance significantly when compared with $x=1$. We can see the correlation improves up to $x=7$. This shows the improvement directly from data augmentation. While there is a computational cost associated with deriving these metrics ($k$ and $x$), we have found these results to be transferable across search spaces eliminating the need to recompute them in the future. All NAS algorithms contain many hyperparameter choices that impact the performance of the algorithm. Often times, these parameters present users with a trade off between computational cost and performance with certain values providing a better ratio of improvement to increased cost. We found that the cost of training $k$ one-shot networks is marginal under the current search space and can easily be expanded with modern compute capabilities to be in the hundreds. The benefits from increasing $k$ plateau at 5 and result in SOAP-NAS being a low cost algorithm. 
% \vspace{-2.5mm}

\section{Conclusion}
% Routability optimization serves a key roll in the modern IC design process. 
% The increase in complexity of modern IC brought a slew of machine learning tools to provide rapid and early DRC hotspot detection. 
DRC hotspot detection poses several challenges for NAS. 
% The change in target metric from a directly trainable one increases the difficulty of search. In addition, the added variance in the ROC-AUC metric further increases difficulty. 
Through smoothing and augmenting, we are able to significantly boost NAS performance and combat these domain-specific issues. 
% SOAP-NAS outperforms existing works by a significant margin achieving 40\% closer to ideal ROC-AUC. 
% Through improving the accuracy of ML models, we are able to significantly improve the capabilities of EDA tools. 
We hope that NAS can continue to improve DRC hotspot detection and a multitude of similar problems in the EDA domain such as optical proximity correction, clock tree prediction, lithography hotspot detection and IR drop estimation.

\bibliographystyle{ieeetr}
\bibliography{soap-nas}

\begin{thebibliography}{10}

\bibitem{routenet}
Z.~Xie, Y.-H. Huang, G.-Q. Fang, H.~Ren, S.-Y. Fang, Y.~Chen, and J.~Hu, ``Routenet: Routability prediction for mixed-size designs using convolutional neural network,'' in {\em 2018 IEEE/ACM International Conference on Computer-Aided Design (ICCAD)}, pp.~1--8, IEEE, 2018.

\bibitem{nas-craft}
J.~Pan, C.~Chang, T.~Zhang, and et~al, ``Automatic routability predictor development using neural architecture search,'' {\em ICCAD'21}, 2020.

\bibitem{drc-cnn}
R.~Liang, H.~Xiang, D.~Pandey, L.~Reddy, S.~Ramji, G.-J. Nam, and J.~Hu, ``Drc hotspot prediction at sub-10nm process nodes using customized convolutional network,'' in {\em ISPD}, pp.~135--142, 09 2020.

\bibitem{pros}
J.~Chen and et~al, ``Pros: A plug-in for routability optimization applied in the state-of-the-art commercial eda tool using deep learning,'' in {\em ICCAD}, ICCAD '20, 2020.

\bibitem{FPGA2_GAN}
M.~B. Alawieh and et~al, ``High-definition routing congestion prediction for large-scale fpgas,'' in {\em 2020 25th Asia and South Pacific Design Automation Conference (ASP-DAC)}, pp.~26--31, IEEE, IEEE Computer Society, 2020.

\bibitem{FPGA3_GAN}
C.~Yu and Z.~Zhang, ``Painting on placement: Forecasting routing congestion using conditional generative adversarial nets,'' in {\em Proceedings of the 56th Annual Design Automation Conference 2019}, pp.~1--6, 2019.

\bibitem{hotspot1}
J.-R. Gao and et~al, ``Lithography hotspot detection and mitigation in nanometer vlsi,'' in {\em 2013 IEEE 10th International Conference on ASIC}, pp.~1--4, IEEE, 2013.

\bibitem{hotspot2}
A.~B. Kahng and et~al, ``Fast dual graph-based hotspot detection,'' in {\em Photomask Technology 2006}, vol.~6349, pp.~125--132, SPIE, 2006.

\bibitem{zhang2024qplacer}
J.~Zhang, H.~Wang, Q.~Ding, J.~Gu, R.~Assouly, W.~D. Oliver, S.~Han, K.~R. Brown, H.~H. Li, and Y.~Chen, ``Qplacer: Frequency-aware component placement for superconducting quantum computers,'' 2024.

\bibitem{cc_sv}
Z.~Qi and et~al, ``Accurate prediction of detailed routing congestion using supervised data learning,'' in {\em 2014 IEEE 32nd international conference on computer design (ICCD)}, pp.~97--103, IEEE, 2014.

\bibitem{cc_sv2}
Z.~Zhou and et~al, ``Supervised-learning congestion predictor for routability-driven global routing,'' in {\em 2019 International Symposium on VLSI Design, Automation and Test (VLSI-DAT)}, pp.~1--4, IEEE, 2019.

\bibitem{FPGA4}
C.-W. Pui and et~al, ``Clock-aware ultrascale fpga placement with machine learning routability prediction,'' in {\em 2017 IEEE/ACM International Conference on Computer-Aided Design (ICCAD)}, pp.~929--936, IEEE, 2017.

\bibitem{pin}
T.-C. Yu and et~al, ``Pin accessibility prediction and optimization with deep learning-based pin pattern recognition,'' in {\em Proceedings of the 56th Annual Design Automation Conference 2019}, pp.~1--6, 2019.

\bibitem{wang2022lhnnlatticehypergraphneural}
B.~Wang, G.~Shen, D.~Li, J.~Hao, W.~Liu, Y.~Huang, H.~Wu, Y.~Lin, G.~Chen, and P.~A. Heng, ``Lhnn: Lattice hypergraph neural network for vlsi congestion prediction,'' 2022.

\bibitem{1e9716e0773d44d2b45ca4038666525b}
S.~Zheng, L.~Zou, P.~Xu, S.~Liu, B.~Yu, and M.~Wong, ``Lay-net: Grafting netlist knowledge on layout-based congestion prediction,'' in {\em 2023 42nd IEEE/ACM International Conference on Computer-Aided Design, ICCAD 2023 - Proceedings}, IEEE/ACM International Conference on Computer-Aided Design, Digest of Technical Papers, ICCAD, (United States), Institute of Electrical and Electronics Engineers Inc., 2023.
\newblock Publisher Copyright: {\textcopyright} 2023 IEEE.; 42nd IEEE/ACM International Conference on Computer-Aided Design, ICCAD 2023 ; Conference date: 28-10-2023 Through 02-11-2023.

\bibitem{FPGA}
A.~Al-Hyari and et~al, ``A deep learning framework to predict routability for fpga circuit placement,'' {\em ACM Transactions on Reconfigurable Technology and Systems (TRETS)}, vol.~14, no.~3, pp.~1--28, 2021.

\bibitem{lissnas}
B.~Gopal, A.~Sridhar, and et~al, ``Lissnas: Locality-based iterative search space shrinkage for neural architecture search,'' {\em IJCAI'23}, 2023.

\bibitem{zoph2017neural}
B.~Zoph and Q.~Le, ``Neural architecture search with reinforcement learning,'' in {\em International Conference on Learning Representations}, 2017.

\bibitem{real2017large}
E.~Real and et~al, ``Large-scale evolution of image classifiers,'' in {\em International Conference on Machine Learning}, pp.~2902--2911, PMLR, 2017.

\bibitem{pmlr-v28-bergstra13}
J.~Bergstra, D.~Yamins, and D.~Cox, ``Making a science of model search: Hyperparameter optimization in hundreds of dimensions for vision architectures,'' in {\em Proceedings of the 30th International Conference on Machine Learning} (S.~Dasgupta and D.~McAllester, eds.), vol.~28 of {\em Proceedings of Machine Learning Research}, (Atlanta, Georgia, USA), pp.~115--123, PMLR, 17--19 Jun 2013.

\bibitem{wen2020neural}
W.~Wen and et~al, ``Neural predictor for neural architecture search,'' in {\em European Conference on Computer Vision}, pp.~660--676, Springer, 2020.

\bibitem{guo2019singlepath}
Z.~Guo and et~al, ``Single path one-shot neural architecture search with uniform sampling,'' {\em CoRR}, vol.~abs/1904.00420, 2019.

\bibitem{ying2019nasbench}
C.~Ying and et~al, ``Nas-bench-101: Towards reproducible neural architecture search,'' {\em CoRR}, vol.~abs/1902.09635, 2019.

\bibitem{zhang2017shufflenet}
X.~Zhang and et~al, ``Shufflenet: An extremely efficient convolutional neural network for mobile devices,'' in {\em Proceedings of the IEEE conference on computer vision and pattern recognition}, pp.~6848--6856, 2018.

\bibitem{transNASBench}
J.~Siems and et~al, ``Nas-bench-301 and the case for surrogate benchmarks for neural architecture search,'' {\em CoRR}, vol.~abs/2008.09777, 2020.

\bibitem{ISCAS}
F.~Brglez and et~al, ``Combinational profiles of sequential benchmark circuits,'' in {\em 1989 IEEE International Symposium on Circuits and Systems (ISCAS)}, pp.~1929--1934, IEEE, 1989.

\bibitem{ITC}
F.~Corno and et~al, ``Rt-level itc'99 benchmarks and first atpg results,'' {\em IEEE Design \& Test of computers}, vol.~17, no.~3, pp.~44--53, 2000.

\bibitem{IWLS}
C.~Albrecht, ``Iwls 2005 benchmarks,'' in {\em International Workshop for Logic Synthesis (IWLS)}, vol.~9, IWLS, 2005.

\bibitem{ISPD}
I.~S. Bustany and et~al, ``Ispd 2015 benchmarks with fence regions and routing blockages for detailed-routing-driven placement,'' in {\em Proceedings of the 2015 Symposium on International Symposium on Physical Design}, pp.~157--164, 2015.

\end{thebibliography}

\end{document}